\documentclass[runningheads]{llncs}

\usepackage[T1]{fontenc}
\usepackage{graphicx,booktabs,tabularx,xcolor,hyperref}
\usepackage{amsmath,subcaption,float,amsfonts,amssymb,bm,mathtools,booktabs,threeparttable,nccmath}
\usepackage[title]{appendix}
\hypersetup{colorlinks=true,linkcolor=[rgb]{0.1,0.3,0.9},urlcolor=[rgb]{0.2,0.2,0.2},citecolor=[rgb]{0.1,0.3,0.9}}

\usepackage{float}

\usepackage{svg}

\begin{document}


\author{Said Djafar Said\inst{1} \and Torkan Gholamalizadeh\inst{2} \and Mostafa Mehdipour Ghazi\inst{1}\thanks{Corresponding author: ghazi@di.ku.dk}}
\authorrunning{S. D. Said et al.}
\institute{Pioneer Centre for AI, Department of Computer Science, University of Copenhagen \and Research and Development, 3Shape A/S, Copenhagen, Denmark}

\title{Tooth-Diffusion: Guided 3D CBCT Synthesis with Fine-Grained Tooth Conditioning}
\titlerunning{Tooth-Diffusion}

\maketitle

\begin{abstract}
 
Despite the growing importance of dental CBCT scans for diagnosis and treatment planning, generating anatomically realistic scans with fine-grained control remains a challenge in medical image synthesis. In this work, we propose a novel conditional diffusion framework for 3D dental volume generation, guided by tooth-level binary attributes that allow precise control over tooth presence and configuration. Our approach integrates wavelet-based denoising diffusion, FiLM conditioning, and masked loss functions to focus learning on relevant anatomical structures. We evaluate the model across diverse tasks, such as tooth addition, removal, and full dentition synthesis, using both paired and distributional similarity metrics. Results show strong fidelity and generalization with low FID scores, robust inpainting performance, and SSIM values above 0.91 even on unseen scans. By enabling realistic, localized modification of dentition without rescanning, this work opens opportunities for surgical planning, patient communication, and targeted data augmentation in dental AI workflows. The codes are available at: \url{https://github.com/djafar1/tooth-diffusion}.

\keywords{CBCT Scan Synthesis \and Tooth Inpainting \and 3D Generative Modeling \and Conditional Diffusion \and FiLM}.

\end{abstract}

\section{Introduction}

Cone-beam computed tomography (CBCT) has become indispensable in dental and maxillofacial imaging, offering high-resolution 3D representations of dentition. However, it remains challenged by inherent limitations such as noise, metal artifacts, and a restricted field of view \cite{scarfe2008cone,pauwels2015technical}. Deep learning methods have demonstrated strong potential in segmentation and reconstruction tasks, yet they often struggle with the anatomical variability of teeth, the presence of missing teeth, and the limited capacity to control or correct for structural artifacts.

Teeth segmentation from CBCT has achieved high accuracy using convolutional neural networks (CNNs), U-Net variants, and attention-based architectures \cite{sadr2024deep,singh2022progress,bolelli2025segmenting}. However, these methods are primarily deterministic and do not support conditional generation for treatment planning, such as simulating missing teeth, implants, bridges, or fillings. Generative models based on generative adversarial networks (GANs) and denoising diffusion probabilistic models (DDPMs) have been explored for image enhancement tasks in CBCT-to-CT and MRI-to-CBCT synthesis \cite{choi2023deep,hu2025cbct,zhang2025texture}, yet prior work on generating synthetic dentition conditioned on user-specified tooth-level attributes remains scarce. To the best of our knowledge, no prior approach enables fine-grained control over individual tooth presence or absence in 3D CBCT, which is a key novelty of this study.

DDPMs have emerged as a robust framework for image synthesis due to their stability and ability to model complex distributions \cite{ho2020denoising}. Recent medical imaging adaptations include CBCT-to-CT translation \cite{hu2025cbct}, limited-angle CBCT reconstruction \cite{gao2025limited}, and medical image denoising \cite{demir2025diffdenoise}. However, these approaches have not been extended to the generation of anatomically accurate, condition-driven dental CBCTs. Such generative capabilities are clinically valuable for simulating anatomical variations, including missing or restored teeth, essential for planning personalized treatments such as implants or orthodontic interventions. Moreover, this can enhance data augmentation, address missing data scenarios, and support the training of robust models in low-resource or imbalanced datasets.

In this work, we propose a novel method to generate synthetic CBCT volumes of dentition with explicit, user-defined tooth configurations. We train a wavelet-based latent diffusion model conditioned on tooth presence, encoded via Feature-wise Linear Modulation (FiLM) embeddings \cite{perez2018film}. By employing a masked L2 loss focused on tooth regions during training, and simulating tooth removal or addition through augmentation, the model achieves precise localization and reconstruction of dental structures. Beyond improving generative controllability, the ability to insert or remove specific teeth enables realistic pre/post-treatment simulations, supporting surgical planning, patient communication, and multidisciplinary case discussion, while also providing a targeted source of variation for augmenting datasets in tasks such as segmentation and detection.

The contributions of this paper are as follows. (1) We introduce an efficient generative framework for guided CBCT dentition synthesis, enabling explicit control over tooth presence at inference time. (2) We incorporate FiLM conditioning and a masked L2 loss to emphasize anatomically realistic reconstruction in tooth regions while suppressing background influence. (3) By simulating tooth removal and addition, we train the model to operate in two distinct modes, completion and removal, enabling scan-aware generation for clinically relevant editing (e.g., implant planning) or robust model training via data augmentation. (4) We conduct comprehensive quantitative and qualitative evaluations, including fairness across tooth positions and fidelity of reconstructed teeth, demonstrating the high realism, variability, and flexibility of the proposed model.

\section{Related Work}

Deep learning has improved automatic tooth segmentation, with convolutional models achieving average Dice scores above 0.9 across maxillary and mandibular scans \cite{polizzi2023tooth}. A meta-analysis of 29 studies confirms these high accuracies and robustness across datasets \cite{sadr2024deep}. However, segmentation remains inherently limited; it does not support image generation or editing, and performance is often degraded by metal artifacts, scanner variability, and background dominance \cite{dot2024dentalsegmentator}.

Despite growing interest in synthetic medical imaging using deep learning models, generative models have been sparsely applied to dental data. Pano-GAN \cite{pedersen2025pano} uses a Wasserstein GAN \cite{arjovsky2017wasserstein} to synthesize 2D panoramic radiographs for augmentation, but it cannot capture full 3D anatomical structure. GANs also suffer from well-documented drawbacks such as mode collapse and difficulty preserving structural consistency, especially in volumetric settings.

DDPMs have emerged as a powerful alternative with greater stability and diversity in image synthesis \cite{ho2020denoising}. They have been applied to CBCT denoising and CT translation, improving quality and downstream tasks like segmentation or dosimetry \cite{hu2025cbct,zhang2025texture}. Recent variants, including DiffDenoise \cite{demir2025diffdenoise} and cycle-consistent diffusion models \cite{gao2025limited}, further enhance reconstruction from sparse or artifact-prone scans. Yet, these methods treat the image volume holistically, lacking mechanisms for localized anatomical control such as tooth editing or generation.

Conditional diffusion frameworks have enabled controllable image synthesis in other domains via latent diffusion modeling (LDM) \cite{rombach2022high}, cross-modal conditioning, and feature-wise transformations like FiLM. While such mechanisms support attribute-driven generation, no existing work addresses conditional 3D CBCT synthesis with anatomically precise, tooth-level guidance.

\section{Methods}

\subsection{Guided Wavelet Diffusion Model}

\begin{figure}[t]
\centering
\includegraphics[width=0.975\textwidth]{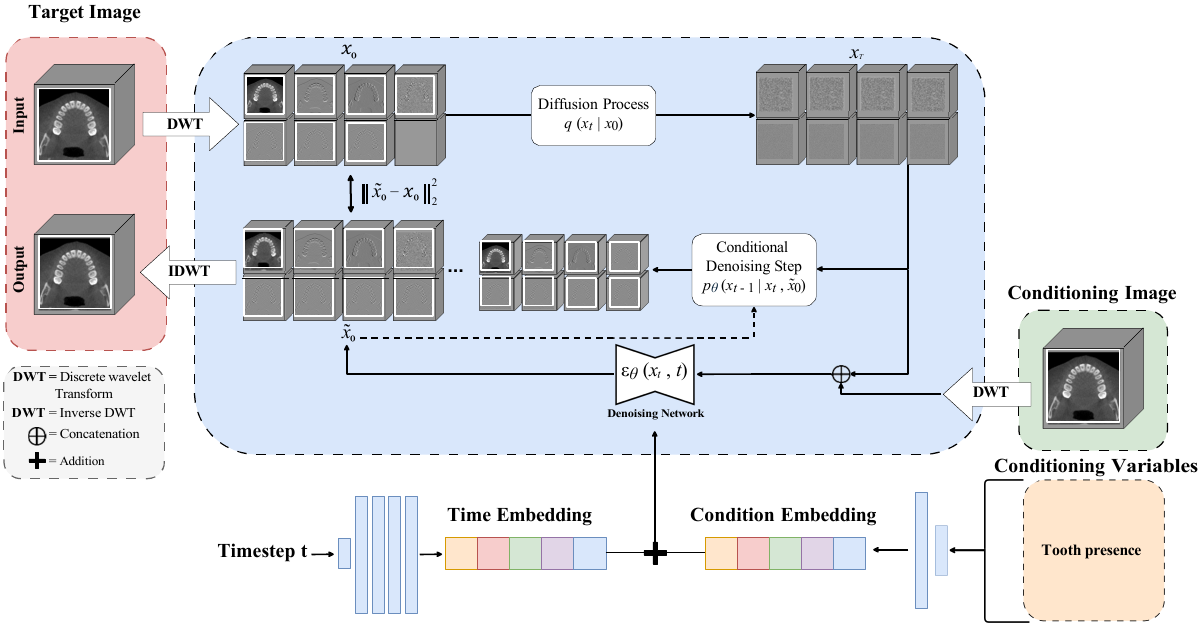}
\caption{Overview of the proposed framework. A guided diffusion model is used for 3D CBCT scan generation with editable tooth configurations.}
\label{fig:diffusion-model}
\end{figure}

We employ a wavelet denoising diffusion model (WDM) \cite{friedrich2024wdm} tailored for 3D CBCT volume synthesis. As illustrated in Fig. \ref{fig:diffusion-model}, the proposed framework follows a conditional generation paradigm, where the model is guided by binary attribute vectors representing tooth presence and a 3D CBCT scan to be edited. This conditional design enables the generation and editing of 3D scans with precise control over dental configurations.

To alleviate the high computational demands and accelerate training and inference, we replace standard Gaussian noise perturbation in pixel space with a wavelet-domain formulation. Specifically, we apply a 3D Haar wavelet transform to decompose the signal into multi-scale frequency components, and inject Gaussian noise into these components. The diffusion model then learns to iteratively denoise in the wavelet domain, operating on representations with half the original spatial resolution. This latent formulation significantly reduces memory and compute requirements while preserving semantic fidelity (e.g., global volume structure) and enhancing detail reconstruction (e.g., tooth boundaries).

\subsection{Condition Embedding}

Each scan is associated with a binary vector of length 32, indicating the presence or absence of individual teeth. This vector is passed through a linear layer to produce a learned conditioning embedding with the same dimensionality as the time embedding, which is obtained using a two-layer multilayer perceptron. These embeddings are combined within each residual block via FiLM \cite{perez2018film}, implemented as a SiLU activation followed by a linear projection, and integrated into the U-Net architecture. FiLM enables the network to modulate intermediate feature activations by applying learned, condition-dependent scaling and shifting, thereby allowing dynamic control based on the desired tooth configuration. During training, the model is conditioned on a real scan and optimized to reconstruct a version consistent with the specified tooth attributes.

\subsection{Tooth Augmentation}

In addition to standard training, where conditions are derived from reconstructing the input scan to match its original dental configuration, we introduce augmentation strategies to promote robustness. Specifically, we simulate two complementary scenarios by modifying the conditioning and target images: tooth addition and tooth removal. In the addition scenario, up to 50\% of the teeth are randomly masked in the conditioning image, while the target remains the original, unaltered scan. The conditioning vector guides the model to plausibly reconstruct the missing structures, and the loss is computed against the original label to penalize inaccurate synthesis. In the removal scenario, up to 50\% of teeth are removed from the target image, while the conditioning image retains the full dental configuration. The model learns to suppress specified regions, with supervision still applied relative to the original, unaltered label. These augmentations simulate clinically relevant use cases, such as handling missing or implanted teeth in surgical planning or restorative workflows.

To ensure that missing teeth appear realistic in the simulated training data, we avoid naive zeroing or masking of the region. Instead, we employ an image-based inpainting strategy to fill the masked tooth cavity with anatomically plausible content. Specifically, we first dilate the tooth mask to accommodate boundary uncertainty and define a broader region for removal. This expanded mask is used to set the corresponding region in the CBCT scan to missing values. Next, we apply the Manhattan (city-block) distance transform to the binary mask, indicating missing regions as zeros. The resulting distance map identifies the nearest valid voxels, whose intensity values are then propagated to fill the cavity. This approach allows the reconstructed region to reflect plausible anatomical structure, such as gradients between air and jawbone intensities near the crown and root, rather than introducing artificial holes to which the model could overfit. Finally, we apply Gaussian smoothing to the inpainted region to eliminate abrupt transitions and promote spatial coherence.

To further increase the effective training sample size, we apply a simple yet effective data augmentation by horizontally flipping the scans and their corresponding label maps (left-to-right). Given the approximate bilateral symmetry of human dentition, we adjust the tooth label values post-flip to preserve anatomical correctness. For the upper jaw (labels 1 to 16), each label is reassigned as $17 - t$, and for the lower jaw (labels 17 to 32), as $49 - t$, where $t$ is the original tooth label. This transformation ensures consistency in left–right orientation and tooth identity, thereby augmenting the dataset without introducing semantic noise.

\subsection{Loss Function}

Given the high background-to-signal ratio in CBCT scans, we introduce a masked L2 loss to concentrate learning on tooth-bearing regions. During training, a soft spatial mask $M$ is derived from the ground truth tooth segmentation by applying a Gaussian blur around tooth boundaries. This mask emphasizes regions near the teeth while down-weighting the less informative background. The masked L2 reconstruction loss is defined as:
\begin{equation}
\mathcal{L}_{\text{Masked}} = \| M \odot (x - \hat{x}) \|_2^2,
\end{equation}
where $x$ is the ground truth scan, $\hat{x}$ is the generated output, $M$ is the soft mask, and $\odot$ denotes element-wise multiplication. This loss penalizes discrepancies, specifically in regions affected by tooth additions or removals, encouraging anatomically faithful reconstructions. The masked loss is then combined with the primary WDM reconstruction loss after the denoising process to yield the total training objective:
\begin{equation}
\mathcal{L}_{\text{Total}} = \mathcal{L}_{\text{WDM}} + \lambda \, \mathcal{L}_{\text{Masked}},
\end{equation}
where $\lambda$ is a weighting factor, empirically set to 10 in our experiments.

\section{Experiments and Results}

\subsection{Data}

We utilize a curated dataset of CBCT scans with ground truth dental segmentation\footnote{\url{https://github.com/ErdanC/Tooth-and-alveolar-bone-segmentation-from-CBCT}} for our study, originally introduced in \cite{cui2019toothnet,cui2021hierarchical,cui2022fully}. From the initial set of 150 CBCT volumes, we exclude 50 scans from a different cohort acquired at higher resolution (0.2 mm$^3$ voxel size) for subsequent analysis. Additionally, we discard two scans with missing segmentation maps and two segmentation volumes without corresponding scans. The resulting 98 volumes are manually relabeled according to the Universal Numbering System (tooth numbers 1--32), excluding supernumerary teeth present in two of the scans. For missing teeth annotations, we provide manual annotations where applicable. Notably, the dataset includes patients with multiple CBCT acquisitions at different treatment stages, enabling analysis of longitudinal consistency and anatomical changes.

All CBCT scans are spatially standardized to a fixed volume size of $256 \times 256 \times 256$ voxels by cropping or padding based on the tooth annotations. This ensures full dentition coverage while preserving the original voxel resolution of 0.4 mm$^3$, and reduces memory usage by excluding excessive background regions. Intensity values are normalized to the $[-1, 1]$ range for stable training. For each scan, binary labels are generated for individual teeth and are used both as supervision during training and for computing the masked reconstruction loss.

\subsection{Experimental Setup}

To scale training across multiple GPUs, we adapted the diffusion model using Distributed Data Parallel framework along with a Distributed Sampler for efficient data loading. The model was trained for 100,000 iterations with 1,000 diffusion time steps using a linear noise schedule. Optimization was performed using Adam optimizer with an initial learning rate of $1 \times 10^{-5}$, scaled linearly with the number of GPUs. A batch size of 1 per GPU was used to accommodate the memory constraints of volumetric data. During evaluation, we test the model on real CBCT scans, both with and without artificially removed/added teeth, and compare reconstructions against the corresponding ground truth volumes.

Out of the 98 available scans, we reserve 8 unique patient scans for testing, selected to represent edge cases such as complete dentition, partial dentition with only a few remaining teeth, or the presence of artifacts like brackets, braces, and mini-screws. The remaining 90 scans are used for training/validation. Although the dataset appears limited in size, each scan encompasses 32 distinct tooth states, creating a combinatorial space of present/missing patterns. Combined with data augmentation during training, this allows diverse testing scenarios in generative settings, where variability rather than sample count is critical.

We evaluate model performance using standard quality metrics for visual fidelity, including the Structural Similarity Index Measure (SSIM) for paired comparisons and the Fréchet Inception Distance (FID) for distributional comparison, both computed in 3D between real and generated volumes. In addition, we assess fairness in generation using per-tooth similarity analysis to identify potential bias in reconstruction fidelity, particularly in scans with a full dental set used for simulated tooth removal and addition scenarios.

\subsection{Results}

\paragraph{\textbf{Reconstruction Synthesis.}} We first assess the model's ability to reconstruct full CBCT scans from conditioning vectors reflecting the original dentition. Quantitative evaluation is conducted using FID between training-validation and training-test splits, providing insight into both overfitting and generalization. The FID score on the test set is notably low (40.27), indicating high-quality image generation relative to the real training samples. It also suggests that the model does not overfit or memorize the training distribution. The relatively higher FID scores for the validation set (88.81) may be attributed to the smaller number of samples (2 vs. 8) used during validation.

\begin{figure}[t]
\centering
\includegraphics[width=0.9\textwidth,height=0.25\textheight]{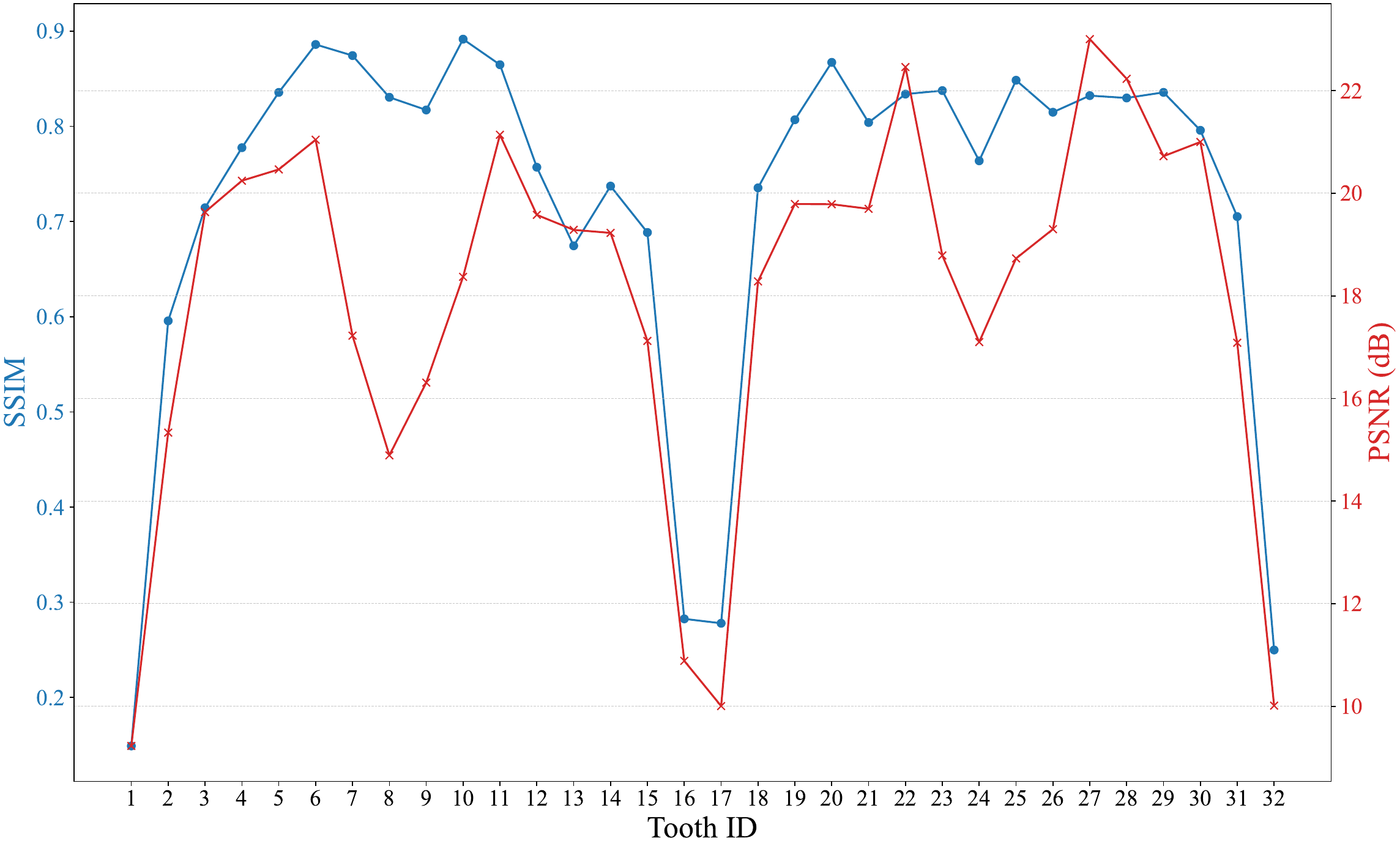}
\caption{Similarities between the original and reconstructed tooth when individually removed and regenerated by the model. Lower similarity is observed for wisdom teeth, likely due to their anatomical variability and data scarcity.}
\label{fig:addition-ssim-bar}
\end{figure}

\paragraph{\textbf{Tooth Addition Synthesis.}} To evaluate single-tooth completion, we simulate missing teeth by masking individual teeth in test scans with complete dentition. The model is then tasked with reconstructing the missing tooth based on the remaining context. We compute the SSIM and PSNR between the reconstructed and ground-truth teeth on a per-tooth basis. The average SSIM scores per tooth are visualized in Fig. \ref{fig:addition-ssim-bar}, highlighting variation across tooth positions. As can be seen, the tooth addition results demonstrate fairly accurate synthesis across most teeth, except for the molars and wisdom teeth (i.e., tooth IDs 1, 16, 17, and 32), which are typically scarce in the training datasets and exhibit greater anatomical variability in size, shape, and orientation.

\paragraph{\textbf{Tooth Removal Synthesis.}} We assess the model's capacity to synthesize realistic scans with specific teeth removed. For this, we define common patterns of tooth absence and apply them to scans with full dentition in the test set. The model generates corresponding scans with these teeth removed. We then compare the generated scans to real samples from the matching tooth-absence groups using FID. Table \ref{tab:removal-fid} reports the FID scores for different target tooth-absence patterns. As shown in the table, the results yield low FID scores across different missing-tooth groups compared to the corresponding generated samples, demonstrating the model's ability to successfully inpaint missing teeth.
 
\begin{table*}[t]
\centering
\small
\caption{Comparison of FID scores between test-time generated scans with removed teeth and training scans exhibiting matching tooth absence.}
\vspace{0.1cm}
\label{tab:removal-fid}
\renewcommand{\arraystretch}{1.15}
\centering
\begin{tabular}{cccc}
\toprule
Missing Teeth \, & [1, 16] \, & [1, 16, 17, 32] \, & [16, 17, 18, 19] \\
\bottomrule
FID Score & 75.20 & 74.36 & 80.03 \\
\toprule
\end{tabular}
\end{table*}

\begin{figure}[t]
\centering
\begin{subfigure}[b]{0.24\linewidth}
\centering
\includegraphics[width=0.99\linewidth]{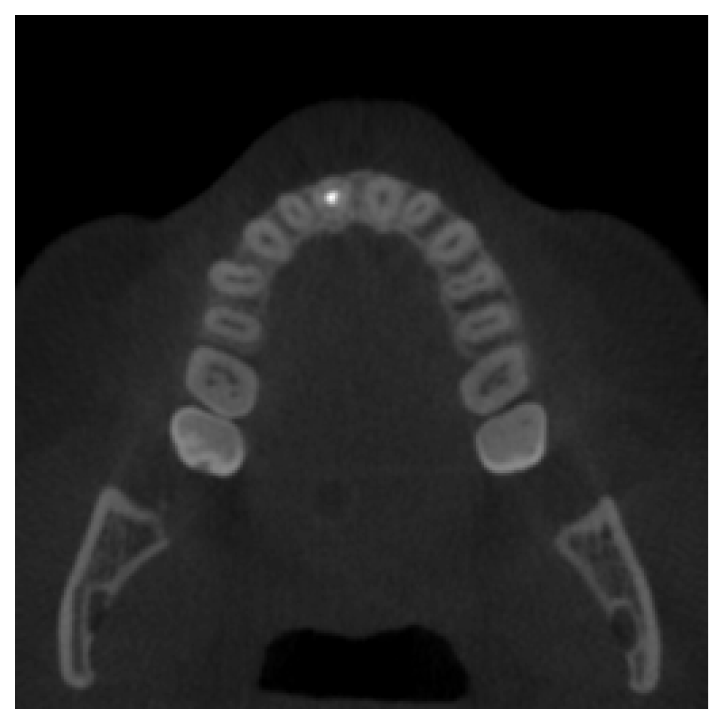}
\caption{Real Scan 1}
\end{subfigure}
\hfill
\begin{subfigure}[b]{0.24\linewidth}
\centering
\includegraphics[width=0.99\linewidth]{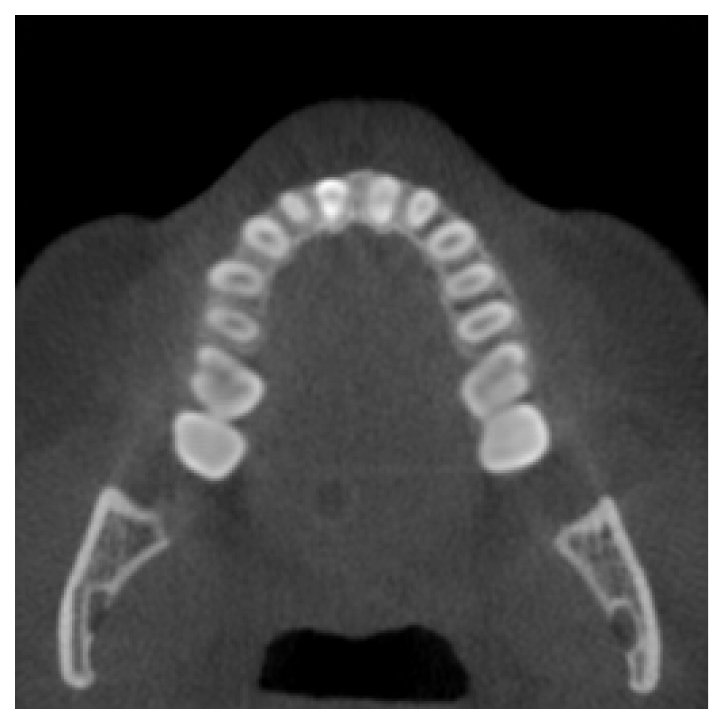}
\caption{Synthetic Scan 1}
\end{subfigure}
\begin{subfigure}[b]{0.24\linewidth}
\centering
\includegraphics[width=0.99\linewidth]{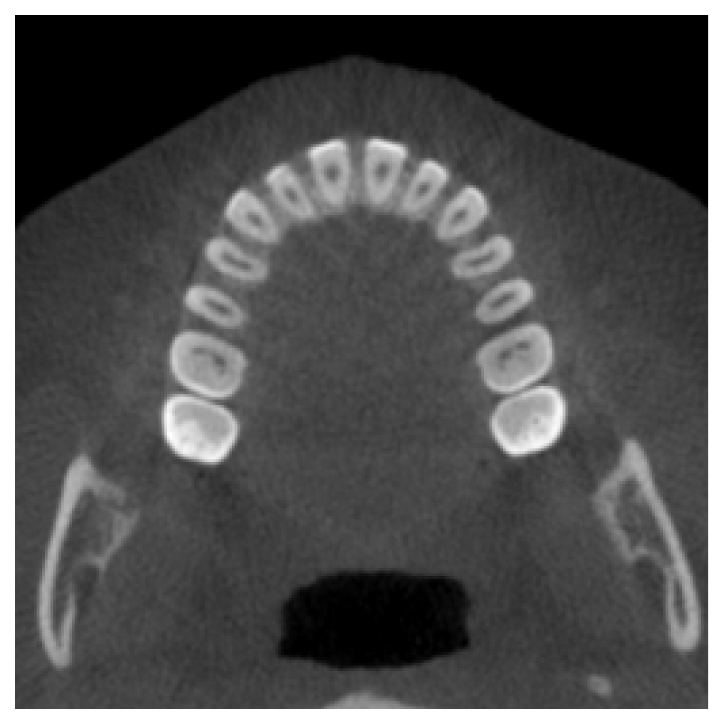}
\caption{Real Scan 2}
\end{subfigure}
\hfill
\begin{subfigure}[b]{0.24\linewidth}
\centering
\includegraphics[width=0.99\linewidth]{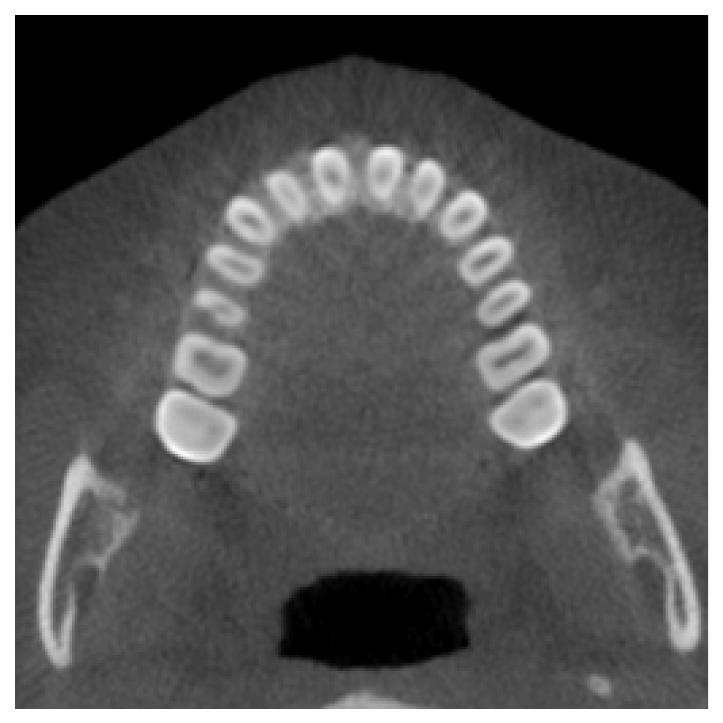}
\caption{Synthetic Scan 2}
\end{subfigure}
\caption{Qualitative comparison between generated CBCT scans and their corresponding real scans with complete dentition.}
\label{fig:full-synthesis}
\end{figure}

\paragraph{\textbf{Full Dental Synthesis.}} As a final experiment, we evaluate the model's performance on generating a complete dentition in scans with no teeth present. This assesses the model's ability to synthesize anatomically plausible full dental structures from the conditioning vector. Fig. \ref{fig:full-synthesis} presents qualitative results comparing the generated samples to real scans with complete dentition. The visual comparison demonstrates a strong alignment between the real and synthetic inpainted regions. Quantitative evaluation supports this observation, with an average SSIM of 0.9123 and an average PSNR of 18.35 computed over the inpainted areas, despite the model not having seen the test samples during training.

\section{Conclusion}

We proposed a guided diffusion framework for controllable synthesis of 3D CBCT dental scans, enabling realistic generation, addition, and removal of teeth based on per-tooth attributes. Our approach integrated a wavelet-based denoising diffusion backbone with FiLM conditioning and masked reconstruction loss alongside tooth augmentations to guide the generative process toward anatomically plausible outputs. Experimental results demonstrated high visual fidelity and generalization performance, with low FID scores in reconstruction and inpainting tasks, and consistent SSIM values across most teeth. The model successfully handles complex cases, such as missing dentition or the presence of artifacts, and shows potential for clinically oriented simulation tasks such as visualizing treatment outcomes or testing AI models under diverse dentition patterns. 

While we focused on per-tooth conditioning, broader clinical factors such as implants, crowns, and bridges remain challenging and represent promising directions for future work. Moreover, although our dataset was limited in size, each scan enabled extensive variability via combinatorial tooth presence patterns, supporting robust generative evaluation. Scaling to larger public datasets, such as ToothFairy\footnote{\url{https://ditto.ing.unimore.it/toothfairy3/}}, and assessing impact on downstream segmentation or detection tasks will be important future steps toward clinical deployment. This study highlights the feasibility of fine-grained, tooth-level controllable generation and provides a tool for simulation, targeted data augmentation, and the development of more customizable and interpretable generative models in dental imaging.

\section*{Acknowledgments}

This project is supported by the Pioneer Centre for AI, funded by the Danish National Research Foundation (grant number P1).



\bibliographystyle{splncs}
\bibliography{references}

\end{document}